\documentclass[runningheads]{llncs}
\usepackage[T1]{fontenc}
\usepackage{graphicx,verbatim}
\usepackage{booktabs}
\usepackage{tabularx}
\usepackage{multirow}
\usepackage{makecell}
\usepackage{hyperref}
\usepackage{bbding}
\begin{document}
\title{TractoRC: A Unified Probabilistic Learning Framework for Joint Tractography Registration and Clustering}
\titlerunning{Unified Framework for Registration and Clustering}
%
\author{Yijie Li\inst{1} \and Xi Zhu\inst{1} \and Junyi Wang\inst{1} \and Ye Wu\inst{2} \and
Lauren J. O'Donnell\inst{3} \and Fan Zhang\inst{1}\textsuperscript{(\Envelope)}}
\authorrunning{Y. Li et al.}
%
\institute{University of Electronic Science and Technology of China, Chengdu, China \and
Nanjing University of Science and Technology, Nanjing, China \and
Harvard Medical School and Brigham and Women's Hospital, Boston. USA\\
\email{fan.zhang@uestc.edu.cn}}

  
\maketitle              
\begin{abstract}
Diffusion MRI tractography enables \textit{in vivo} reconstruction of white matter (WM) pathways. Two key tasks in tractography analysis include: 1) \textit{tractogram registration} that aligns streamlines across individuals, and 2) \textit{streamline clustering} that groups streamlines into compact fiber bundles. Although both tasks share the goal of capturing geometrically similar structures to characterize consistent WM organization, they are typically performed independently. In this work, we propose \textit{TractoRC}, a unified probabilistic framework that jointly performs tractogram registration and streamline clustering within a single optimization scheme, enabling the two tasks to leverage complementary information. TractoRC learns a latent embedding space for streamline points, which serves as a shared representation for both tasks. Within this space, both tasks are formulated as probabilistic inference over structural representations: registration learns the distribution of anatomical landmarks as probabilistic keypoints to align tractograms across subjects, and clustering learns streamline structural prototypes that capture geometric similarity to form coherent streamline clusters. To support effective learning of this shared space, we introduce a transformation-equivariant self-supervised strategy to learn geometry-aware and transformation-invariant embeddings. Experiments demonstrate that jointly optimizing registration and clustering significantly improves performance in both tasks over state-of-the-art methods that treat them independently. Code will be made publicly available at \url{https://github.com/yishengpoxiao/TractoRC}.

\end{abstract}
\section{Introduction}
\label{sec:introduction}
Diffusion MRI (dMRI) tractography enables \textit{in vivo} reconstruction of white matter (WM) pathways and serves as a fundamental tool for studying brain structural connectivity~\cite{basser2000vivo,zhang2026cross}. In tractography analysis, there are two important computational tasks: 1) \textit{tractogram registration}, which aligns anatomical fiber tract structures across subjects, and 2) \textit{streamline clustering}, which groups streamlines into anatomically meaningful fiber bundles~\cite{zhang2022quantitative}. One common feature of the two tasks is to capture common geometrically similar streamline structures to characterize consistent WM organization. Currently, these tasks are typically performed independently, while the joint analysis to leverage complementary information of the two tasks is lacking.

Tractogram registration methods include volume-based and streamline-based approaches. Volume-based methods align diffusion-derived images, such as fractional anisotropy (FA), using conventional image registration frameworks, but often fail to preserve fine-grained streamline geometry~\cite{moulton2018comparison,zhang2021deep}. Streamline-based methods instead directly optimize geometric distances between streamlines, typically requiring iterative correspondence estimation~\cite{garyfallidis2015robust,o2012unbiased}. Recent keypoint-based methods improve robustness but remain limited in capturing globally consistent anatomical alignment across subjects~\cite{wang2025novel}. In parallel, streamline clustering methods aim to group streamlines into coherent bundles based on geometric similarity. Classical approaches rely on pairwise similarity computation in the original space, leading to high computational cost and limited scalability~\cite{garyfallidis2012quickbundles,o2007automatic}. More recent learning-based methods perform clustering in latent spaces to improve efficiency and robustness~\cite{chen2023deep}. However, they do not explicitly incorporate cross-subject alignment, resulting in potentially inconsistent cluster assignments.

Recent advances in point cloud registration offer an alternative by modeling geometric structures as distributions over representative components~\cite{eckart2018hgmr,liao2022fuzzy}. Instead of enforcing explicit point-to-point correspondences, these approaches represent each point cloud using a set of latent components defined by centers and membership weights, and perform registration via distribution matching. By modeling structures at the distributional level, these methods enable robust alignment under structural heterogeneity.

Building on this insight, we propose \textit{TractoRC}, a unified probabilistic framework that jointly performs tractogram registration and streamline clustering within a single optimization scheme. By coupling the two tasks, the framework enables clustering-guided alignment and registration-refined clustering, allowing them to mutually reinforce each other. Unlike prior approaches that treat these tasks independently, our method unifies them within a shared representation space. Our contributions are threefold: 1) a unified probabilistic formulation for joint registration and clustering; 2) a transformation-equivariant self-supervised pretraining strategy for learning geometry-aware and transformation-invariant embeddings; and 3) a joint optimization scheme that enables mutual reinforcement between registration and clustering.

\section{Method}
\label{sec:method}
\begin{figure}[t]
\centering
\includegraphics[width=\textwidth]{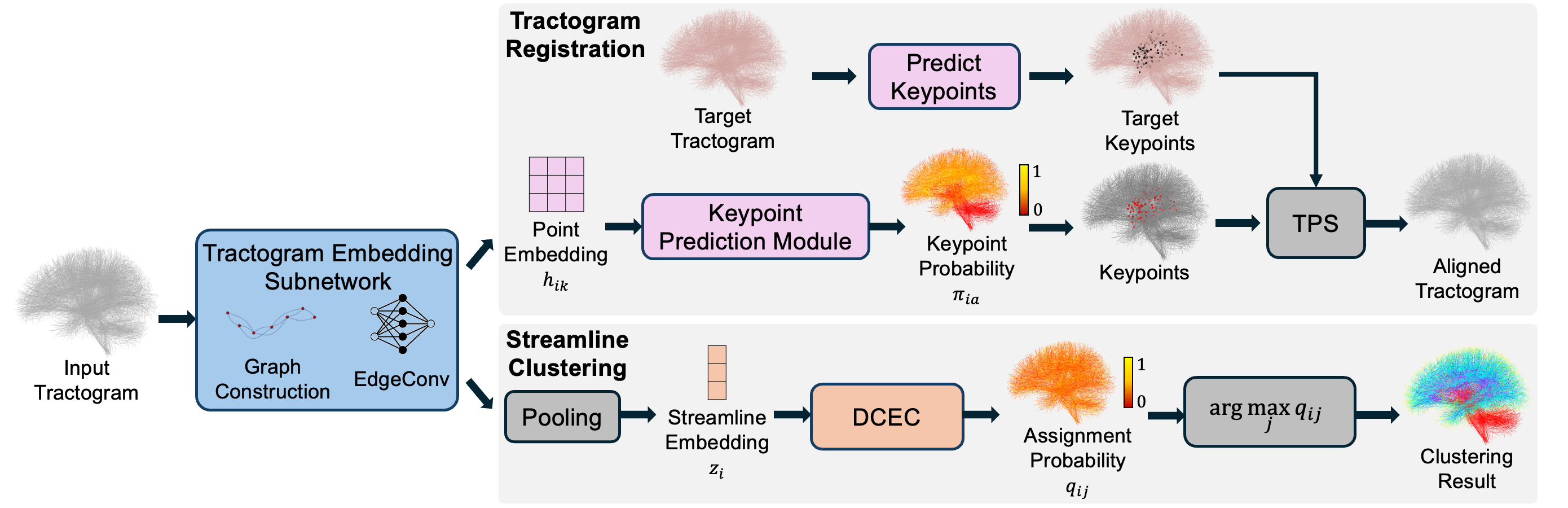}
\caption{Overview of the proposed TractoRC framework.} \label{fig:method}
\end{figure}

Fig.~\ref{fig:method} illustrates the TractoRC framework for joint tractogram registration and streamline clustering. First, a tractogram embedding network maps each streamline point to a latent feature space. These embeddings are then fed into two parallel subnetworks for registration and clustering. In the registration branch, a keypoint prediction module extracts representative streamline keypoints capturing shared anatomical landmarks across subjects, followed by a thin-plate spline (TPS) transformation~\cite{rohr2002landmark,zhao2022thin} to align the input tractogram to the target. In the clustering branch, the point embeddings are aggregated into streamline-level embeddings and fed into a Deep Convolutional Embedded Clustering (DCEC) model~\cite{chen2023deep,guo2017deep} to cluster streamlines into anatomically meaningful bundles. To ensure robust initialization, we perform self-supervised pretraining of the streamline embedding network that learns geometry-aware and transformation-invariant representations, enabling discriminative modeling of streamline structures. The pretrained weights are then used to initialize the full network, after which the registration and clustering subnetworks are jointly optimized, enabling mutual reinforcement between the clustering and registration tasks.
\subsection{Network Architecture}
\label{subsec:network}
\subsubsection{Tractogram Embedding Subnetwork.}The tractogram embedding subnetwork computes point-wise embeddings for each streamline point. In our work, tractograms are represented as point clouds, where each streamline consists of a sequence of 3D points. To capture the contextual information provided by the neighborhood relationships among points, we adopt the Dynamic Graph Convolutional Neural Network (DGCNN) model~\cite{chen2023deep,wang2019dynamic}. Specifically, the model takes the point spatial coordinates as input. For a streamline with $N_p$ sampled points, a $k$-nearest neighbor graph is dynamically constructed and processed by EdgeConv layers to capture local geometric structures. The outputs are point-wise embeddings $h_{ik}$ for the $k$-th point of the $i$-th streamline.
\subsubsection{Tractogram Registration Subnetwork.}The registration subnetwork first estimates a set of keypoints from the computed point-wise embeddings for the input tractogram and a target tractogram randomly selected from the training dataset. To do so, we model keypoints as expectations over a learned spatial distribution~\cite{wang2023robust,wang2025novel}, with the $a$-th keypoint defined as $p_a=\sum_{i}{\pi_{ia}x_i}$, where $x_i$ denotes the $i$-th point and $\pi_{ia}$ is the probability of assigning $x_i$ to keypoint $a$, predicted by a learnable classifier (1D convolutions). This constrains keypoints to lie within the input support, avoids unstable coordinate regression, and provides compact structural components for distributional alignment across subjects.

Then, the predicted keypoints serve as control points for registration between the source and target tractograms. Correspondence is implicitly established by the shared keypoint prediction module via consistent keypoint ordering, and a TPS transformation is estimated in closed form~\cite{rohr2002landmark,zhao2022thin}:
\begin{equation}
    R(x)=Ax+U(x)W,
\label{eq:tps}
\end{equation}
where $x$ denotes a spatial point in the source tractogram, $A$ models the global affine transformation, $W$ captures local deformation, and $U(x)=x^2\ln{x}$ encodes the non-linear relationships between $x$ and the control keypoints. The warped source and target tractograms are then used for registration loss computation (see Section~\ref{subsubsec:optimization}).

\subsubsection{Streamline Clustering Subnetwork.}The clustering subnetwork first computes streamline-level embeddings $z_i$ by aggregating point-wise embeddings along each streamline via symmetric pooling. These streamline embeddings are then fed into a DCEC model for clustering~\cite{chen2023deep,guo2017deep}. With learnable centroids $\mu_j$, soft assignments are computed using Student's $t$-distribution:
\begin{equation}
\label{eq:tstudent}
    q_{ij}=\frac{(1+\Vert z_i-\mu_j\Vert^2)^{-1}}{\sum_{j'}(1+\Vert z_i-\mu_j\Vert^2)^{-1}},
\end{equation}
where $q_{ij}$ denotes the probability of assigning $z_i$ to centroid $\mu_j$. The embeddings and centroids are jointly refined, enabling adaptive feature learning and clustering. The centroids can be interpreted as representative structures, where each streamline is assigned to them via $q_{ij}$, modeling tractography as a distribution over latent components.
\subsection{Joint Registration-Clustering Model Training}
\label{subsec:joint_train}
\subsubsection{Self-Supervised Pretraining.}\label{subsubsec:pretrain}
To learn stable and geometry-aware representations for the tractogram embedding subnetwork, we introduce a self-supervised pretraining scheme based on transformation equivariance. Given a tractogram $T$, an augmented sample is generated via a random affine transformation $G$ and a nonlinear deformation $\Phi$.

For the keypoint prediction module $f(\cdot)$, we enforce keypoint transformation equivariance by minimizing $\Vert f(G\cdot\Phi(T))-(G\cdot\phi)\odot f(T)\Vert$, ensuring that predicted keypoints transform consistently under spatial transformations. A diversity regularization term is further introduced to encourage spatial dispersion among keypoints by penalizing pairwise proximity, preventing keypoint collapse.

In addition, we enforce geometric consistency of streamline embeddings by aligning embedding distances with streamline similarity. Let $z_i$ and $z_j$ denote the embeddings of streamlines $i$ and $j$, and $d(i, j)$ their Minimum Average Direct-Flip (MDF) distance~\cite{garyfallidis2012quickbundles,zhang2018anatomically}, we define a metric alignment loss $\sum_{i,j}Huber(\Vert z_i-z_j\Vert_2-d(i,j))$, encouraging embedding distances to reflect geometric similarity.

The overall pretraining loss averages these two objectives, promoting embeddings that capture both geometric similarity for clustering and transformation consistency for registration.

\subsubsection{Joint Optimization.}\label{subsubsec:optimization}
For training the overall network, the pretrained weights are first used to initialize the tractogram embedding subnetwork. The model is subsequently jointly optimized using randomly sampled tractogram pairs, denoted as the input $T_I$ and target $T_T$. The training objective consists of two complementary components.

For the registration, let $R(\cdot)$ denote the TPS transformation applied to the input tractogram. The loss is defined as a symmetric nearest-neighbor distance:
\begin{equation}
\label{eq:regloss}
    \frac{1}{N_I}\sum_i\min_jd(R(s_i^I),s_j^T)+\frac{1}{N_T}\sum_j\min_id(R(s_i^I),s_j^T)
\end{equation}
where $s^I_i,s^T_j$ are streamlines from the input and target tractogram, and $N_I$ and $N_T$ denote their number of streamlines. This formulation enforces bidirectional geometric consistency between the two tractograms.

For clustering, k-means is first applied to all streamline-level embeddings to initialize the DCEC centroids. We then retain the metric alignment constraint and refine cluster assignments using a Kullback-Leibler divergence loss~\cite{guo2017deep}. Given the soft assignment probability $q_{ij}$, the target distribution is defined as $p_{ij}=\frac{q_{ij}^2/\sum_i q_{ij}}{\sum_{j'}(q_{ij'}^2/\sum_i q_{ij'})}$, which emphasizes high-confidence assignments. The clustering loss is $\sum_{i,j}p_{ij}\log\frac{p_{ij}}{q_{ij}}$, encouraging more discriminative clustering while preserving structural consistency in the embedding space.

The final loss is the average of the above terms, jointly optimizing registration and clustering to enable mutual refinement between geometric alignment and latent structural representations.
\subsection{Model Inference}
\subsubsection{Tractogram Registration.}During inference, tractogram registration is performed within a unified keypoint-based framework. Given input and target tractograms, subject-specific keypoints are first predicted via the keypoint prediction module. The TPS transformation $R(\cdot)$ is then estimated from these keypoints, and applied to warp the input tractogram into the target space according to Eq.~\ref{eq:tps}. As the transformation estimation relies solely on keypoints, the method avoids iterative streamline matching and remains efficient and scalable.
\subsubsection{Streamline Clustering:}For streamline clustering, soft assignments $q_{ij}$ are computed using Eq.~\ref{eq:tstudent}, and each streamline is assigned to the cluster with the highest probability $y_i=\arg\max_j q_{ij}$. To improve robustness, we define the assignment confidence as $q_i^{\max}$ and discard streamlines with low-confidence assignments below a predefined threshold $thr=0.4$.
\subsection{Implementation Details}
The proposed framework is implemented in PyTorch 2.7.1 and trained on NVIDIA RTX 4090 GPUs. We use 128 keypoints and 800 clusters, and resampled each streamline to $N_p=14$ points~\cite{chen2023deep,wang2023robust,zhang2018anatomically}. For DGCNN, a $k$-nearest neighbor graph with $k = 4$ is constructed following prior studies~\cite{chen2023deep}. Pretraining is conducted for 4000 epochs using AdamW with an initial learning rate of $10^{-3}$, decayed by 0.1 every 1000 epochs, followed by 10 epochs of joint training at $10^{-4}$.
\section{Experimental and Results}
\label{sec:experimet}
\begin{table}[t]
\caption{Ablation study on registration and clustering. Bold: best.}
\label{tab:ablation}
\centering
\begin{tabularx}{\linewidth}{
l
>{\centering\arraybackslash}X
>{\centering\arraybackslash}X
>{\centering\arraybackslash}X
>{\centering\arraybackslash}X
}
\toprule
\textbf{Variants} & \multicolumn{2}{c}{\textbf{Registration}} 
& \multicolumn{2}{c}{\textbf{Clustering}} \\
\cmidrule(lr){2-3} \cmidrule(lr){4-5}
& \textbf{ABD} $\downarrow$ 
& \textbf{wDice} $\uparrow$ 
& $\mathbf{\alpha}$ $\downarrow$
& \textbf{WMPG} $\uparrow$ \\
\midrule
\makecell[l]{w/o pretraining \\ \& w/o clustering} & $3.24 \pm 0.69$ & $63.37\% \pm 8.99\%$ & -- & -- \\
\midrule
w/o clustering & $3.12 \pm 0.66$ & $65.04\% \pm 8.69\%$ & -- & -- \\
\midrule
w/o registration & -- & -- & $7.93 \pm 3.06$ & $99.52\% \pm 0.42\%$ \\
\midrule
TractoRC & $\mathbf{2.90 \pm 0.61}$ & $\mathbf{68.26\% \pm 8.71\%}$ & $\mathbf{7.77 \pm 3.18}$ & $\mathbf{99.88\% \pm 0.38\%}$ \\
\bottomrule
\end{tabularx}
\end{table}

\subsection{Dataset and Preprocessing}
Experiments were conducted on the Human Connectome Project Young Adult (HCP-YA) dMRI dataset~\cite{van2013wu} with generalized q-sampling imaging (GQI) reconstruction~\cite{yeh2010generalized} from FiberDataHub~\cite{yeh2025dsi} to compute tractograms. All subjects were linearly registered to MNI space before training. In total, 140 subjects were used and split into training, validation, and testing sets with a ratio of 100:20:20. Whole-brain tractography was generated using DSI Studio with default parameters and the RK4 algorithm~\cite{yeh2013deterministic}. For efficiency, 100,000 streamlines per subject were randomly sampled for training and validation, while 500,000 streamlines were retained for testing.
\subsection{Ablation Study}
We conducted an ablation study to examine the mutual benefits between tractogram registration and streamline clustering.

For registration, we consider three variants: 1) w/o pretraining \& w/o clustering, 2) w/o clustering, and 3) the full TractoRC framework. Performance was evaluated on anatomical tracts extracted via ROI-based segmentation, using the weighted Dice score (wDice)~\cite{cousineau2017test,zhang2019test} and the Average Bundle Distance (ABD)~\cite{garyfallidis2015robust}, which measure overlap and geometric consistency, respectively. As shown in Tab.~\ref{tab:ablation}, each component consistently improves performance: pretraining enhances geometric stability and reduces sensitivity to initialization, while clustering further improves alignment accuracy by providing structural priors.

For clustering, the centroids in DCEC are initialized based on pretrained embeddings, which capture the underlying geometric structures, making the pretraining stage essential for stable initialization. Therefore, in the ablation study, we retain the pretrained embedding network and remove only the Tractogram Registration Subnetwork, resulting in a model equivalent to DeepFiberClustering (DFC)~\cite{chen2023deep}, a deep learning-based unsupervised clustering method. Clustering quality was evaluated using the $\alpha$ measure, which evaluates intra-cluster compactness~\cite{vazquez2020ffclust}, and White Matter Parcellation Generalization (WMPG), which measures cross-subject reproducibility~\cite{o2007automatic}. As shown in Tab.~\ref{tab:ablation}, the proposed method outperforms DFC on both metrics, indicating that registration improves cross-subject consistency and produces more compact clusters.

Overall, the ablation study results show the benefits of the joint formulation, where clustering and registration mutually reinforce each other.
\subsection{Tractogram Registration Compared with SOTA Methods}
\begin{table}[t]
\caption{Comparison of registration performance. Bold: best; underline: second-best.}
\label{tab:reg_results}
\centering
\begin{tabularx}{\linewidth}{>{\raggedright\arraybackslash}X>{\raggedright\arraybackslash}X>{\centering\arraybackslash}X>{\centering\arraybackslash}X}
\toprule
\multicolumn{2}{l}{\textbf{Methods}} & \textbf{ABD} $\downarrow$ & \textbf{wDice} $\uparrow$ \\
\midrule
\multirow{2}{*}{Volume-based} & SyN & \underline{$3.09 \pm 0.63$} & $\mathbf{70.02\% \pm 7.54\%}$ \\
 & SynthMorph & $4.19 \pm 1.21$ & $57.41\% \pm 13.61\%$ \\
\midrule
\multirow{2}{*}{Streamline-based} & WMA & $3.23 \pm 0.79$ & $64.71\% \pm 11.34\%$ \\
 & TractoRC & $\mathbf{2.90 \pm 0.61}$ & \underline{$68.26\% \pm 8.71\%$} \\
\bottomrule
\end{tabularx}
\end{table}
\begin{figure}[t]
\centering
\includegraphics[width=\textwidth]{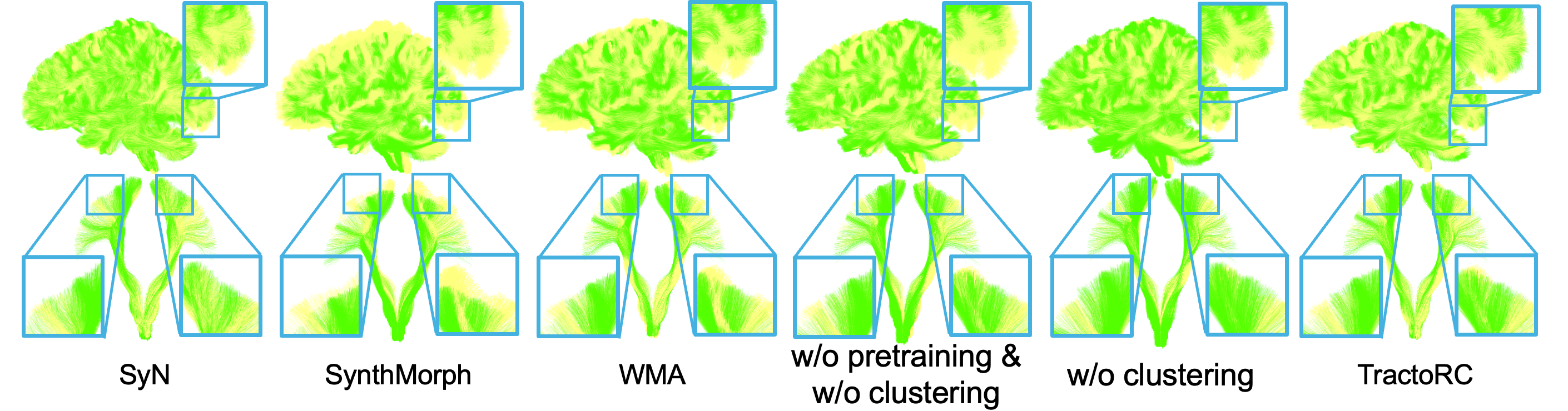}
\caption{Visualization of registration performance across diﬀerent methods. The aligned tractogram is shown in green, and the target tractogram is shown in yellow.} \label{fig:registration_result}
\end{figure}
\begin{table}[t]
\caption{Comparison of clustering performance. Bold: best; underline: second-best.}
\label{tab:clustering_result}
\centering
\begin{tabularx}{\linewidth}{l>{\centering\arraybackslash}X>{\centering\arraybackslash}X}
    \toprule
    \textbf{Methods} & $\mathbf{\alpha}$ $\downarrow$ & \textbf{WMPG} $\uparrow$ \\
    \midrule
WMA & $9.25 \pm 4.17$ & $97.49\% \pm 2.55\%$ \\
QB & $8.43 \pm 2.41$ & $\mathbf{100\% \pm 0\%}$ \\
DFC & \underline{$7.93 \pm 3.06$} & $99.52\% \pm 0.42\%$ \\
TractoRC & $\mathbf{7.77 \pm 3.18}$ & \underline{$99.88\% \pm 0.38\%$} \\
    \bottomrule
\end{tabularx}
\end{table}
We compared the proposed method with both volume- and streamline-based registration approaches. Volume-based baselines include SyN (ANTs)~\cite{avants2008symmetric} and SynthMorph (FreeSurfer)~\cite{hoffmann2021synthmorph}, where transformations were estimated from FA images and applied to the tractogram using 3D Slicer~\cite{fedorov20123d,zhang2020slicerdmri}. Streamline-based baseline is the nonlinear registration framework in White Matter Analysis (WMA)~\cite{o2012unbiased}. Registration performance was evaluated using wDice and ABD.

As shown in Tab.~\ref{tab:reg_results}, the proposed method achieves the best ABD, with all pairwise t-tests yielding $p < 0.01$, indicating improved geometric consistency. In terms of wDice, our method outperforms all competing approaches ($p < 0.01$) except SyN, which achieves the highest wDice, likely due to its volumetric formulation favoring voxel-wise overlap. Qualitative results (Fig.~\ref{fig:registration_result}) further demonstrate more coherent tract alignment and improved structural consistency.

\subsection{Clustering Results Compared with SOTA Methods}
We compare the proposed method against three representative streamline clustering approaches: WMA~\cite{o2013fiber,o2017automated}, QuickBundles (QB)~\cite{garyfallidis2012quickbundles}, and DFC~\cite{chen2023deep}. WMA and DFC are atlas-based methods, while QB performs subject-wise clustering. Clustering quality was evaluated using the $\alpha$ measure and WMPG.
\begin{figure}[t]
\centering
\includegraphics[width=0.95\textwidth]{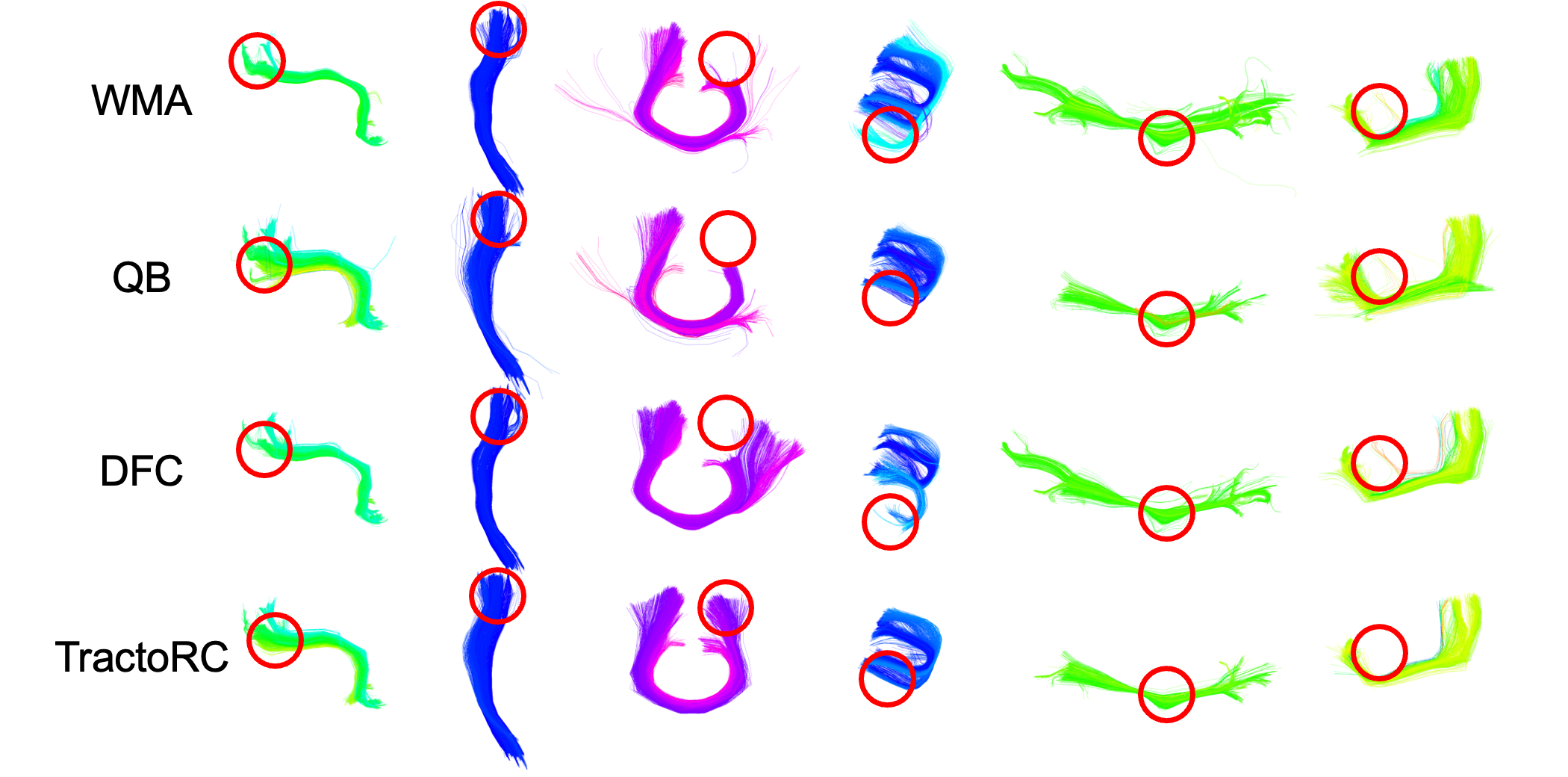}
\caption{Visualization of clustering performance across diﬀerent methods.} \label{fig:clustering_result}
\end{figure}

As shown in Tab.~\ref{tab:clustering_result}, the proposed method improves performance on both metrics, with statistically significant gains over competing methods ($p < 0.01$), except for WMPG in QB. QB achieves the highest WMPG due to its subject-wise clustering strategy, while other methods enforce a shared clustering structure across subjects. Visual comparisons (Fig.~\ref{fig:clustering_result}) further demonstrate that our method produces more spatially coherent fiber bundles.

\section{Conclusion}
This paper presents a unified probabilistic framework for joint tractogram registration and streamline clustering. By integrating structural embeddings, probabilistic keypoints prediction, differentiable transformations, and embedding-based clustering, the method enables clustering-guided alignment and registration-refined clustering within a single optimization scheme. Experiments demonstrate improved alignment accuracy and clustering consistency.

\begin{credits}
\subsubsection{\ackname}This work is in part supported by the National Key R\&D Program of China (No. 2023YFE0118600) and the National Natural Science Foundation of China (No. 62371107).  

\subsubsection{\discintname}The authors have no conflicts of interest to declare.
\end{credits}

    
%
%
%
\bibliographystyle{splncs04}
\bibliography{mybibliography}
\end{document}